\documentclass{article}
\usepackage{spconf,amsmath,epsfig}
\usepackage{amssymb}
\usepackage{multirow}

\title{ben-ge: Extending BigEarthNet with Geographical and Environmental Data}
\name{Michael Mommert$^\mathrm{1}$, Nicolas Kesseli$^\mathrm{1}$, Jo\"{e}lle Hanna$^\mathrm{1}$, Linus Scheibenreif$^\mathrm{1}$, Damian Borth$^\mathrm{1}$, Beg\"{u}m Demir$^\mathrm{2}$}
\address{(1): School of Computer Science, Universit\"{a}t St.~Gallen; (2): Technische Universit\"{a}t Berlin}
\begin{document}
%
\maketitle
\begin{abstract}
Deep learning methods have proven to be a powerful tool in the analysis of large amounts of complex Earth observation data. However, while Earth observation data are multi-modal in most cases, only single or few modalities are typically considered.
In this work, we present the \emph{ben-ge} dataset, which supplements the \emph{BigEarthNet-MM} dataset by compiling freely and globally available geographical and environmental data. Based on this dataset, we showcase the value 
of combining different data modalities for the downstream tasks of patch-based land-use/land-cover classification and land-use/land-cover segmentation. \emph{ben-ge} is freely available and expected to serve as a test bed for 
fully supervised and self-supervised Earth observation applications.
\end{abstract}

\section{Introduction}
\label{sec:intro}

The amount of Earth observation data grows at an ever-increasing rate. To cope with the vast amount and the complexity of the data, scalable and flexible methods are required for their systematic analysis. 
End-to-end Deep Learning approaches have proven highly successful in extracting insights from such complex data across a variety of downstream tasks. 
Furthermore, data fusion, the combination of different data modalities of the observed scene, is beneficial for most applications as it provides additional, and in many cases complementary, information on the scene. 

Most Deep Learning applications for Earth observation rely on supervised learning approaches that require (large amounts of) annotated data, which are typically expensive and tedious to acquire. 
Self-supervised learning approaches, which do not require annotated data, have shown the ability to successfully pretrain deep learning models, which in turn require a significantly smaller amount of 
annotated data for a given downstream task while at the same time outperforming fully supervised approaches \cite{Scheibenreif2022ISPRS, Scheibenreif2022Earthvision}. 
In combination with data fusion across multiple data modalities, richer representations of the underlying data can be learned, further improving the trained model performance and data efficiency.

In order to evaluate the impact of combining different data modalities on fully-supervised and self-supervised learning approaches for Earth observation applications, a dedicated dataset is needed. 
Currently available large-scale Earth observation datasets \cite{Sumbul2019, Sumbul2021, Schmitt2019, Wang2022} comprise multiple data modalities, but those are typically limited to multispectral and synthetic aperture radar (SAR) data. 
However, other data modalities such as meteorological conditions at the time of observation, the topography of the scene or other geographic features are likely to 
support the learning process and improve the performance of the Deep Learning model.

In this work, we present \emph{ben-ge}, an extension to the \emph{BigEarthNet-MM} dataset \cite{Sumbul2021}, in which we supplement the already existing multispectral imaging (Sentinel-2) and SAR polarization (Sentinel-1) data by adding freely and globally 
available data modalities related to geography and environmental conditions. This extension will enable researchers to readily experiment with a wide variety of data modalities for a range of downstream tasks and use cases. 



\section{The \emph{ben-ge} Dataset}
\label{sec:data}

\begin{figure}[t!]
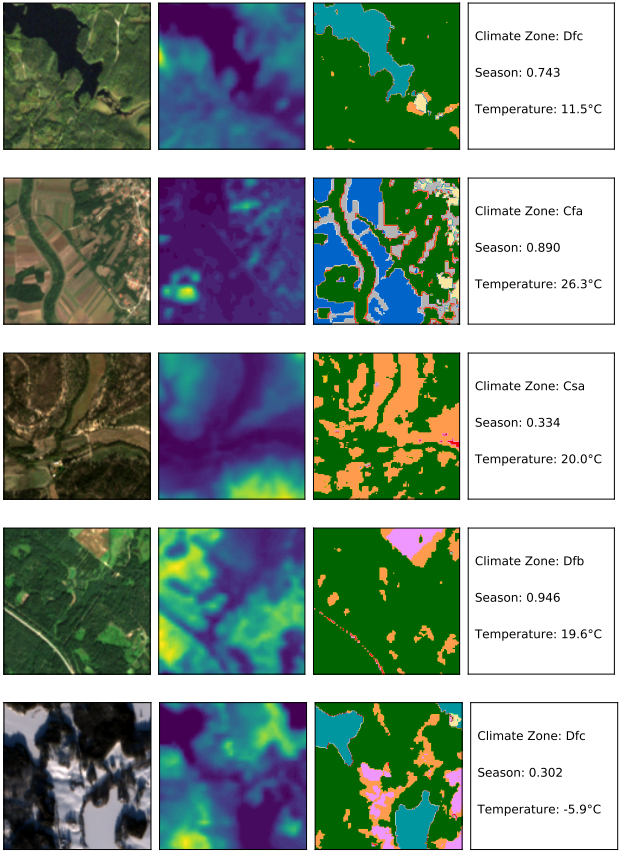

  \centering
  \includegraphics[width=\linewidth]{plots/sample\_000001.pdf}
  \includegraphics[width=\linewidth]{plots/sample\_145253.pdf}
  \includegraphics[width=\linewidth]{plots/sample\_145254.pdf}
  \includegraphics[width=\linewidth]{plots/sample\_032645.pdf}
  \includegraphics[width=\linewidth]{plots/sample\_345243.pdf}
  \caption{\emph{BigEarthNet} and \emph{ben-ge} dataset samples. Each row shows different data modalities of the same scene; columns are (from left to right): Sentinel-2 RGB image (\emph{BigEarthNet}), elevation, LULC and other patch-specific data (see text for details).}
  \label{fig:res}
\end{figure}

\emph{ben-ge} supplements each of the 590,326 \emph{BigEarthNet} patches with freely available geographic and environmental data. Geographic data is provided in the form of patch-based climate-zone classifications, topographic maps, as well as land-use/land-cover maps, while environmental data
is provided in the form of the season and meteorological data concurrent with the Sentinel-1 and Sentinel-2 observations. 
Patch-based \textbf{climate-zone classifications}, following the K\"{o}ppen-Geiger classification scheme, were extracted from \cite{Beck2018}.
\textbf{Topographic maps} are generated based on the Copernicus Digital Elevation Model (GLO-30) \cite{DEM} and interpolated (bilinear resampling) to 10~m resolution on the ground. 
We also provide cropped \textbf{land-use/land-cover (LULC) classification maps} for each \emph{BigEarthNet} patch from the ESA WorldCover 10~m 2021 dataset \cite{Zanaga2022}. As part of this work, we use these LULC maps as targets for model benchmarking (see Section \ref{sec:experiments}), but they could also be used as additional input modality. 
We encode the \textbf{season} at the time of observation (separately for Sentinel-1/2 data) on a range from zero (winter solstice) to unity (summer solstice) based on a sinusoidal projection of the day of the year at the time of observation.
\textbf{Weather data} at the time of observation (temperature at 2~m above the ground, relative humidity, wind vectors at 10~m above the ground) are extracted from the 
ERA-5 global reanalysis \cite{Hersbach2020} for the pressure level at the mean elevation of the observed scene and the time of observation (separately for Sentinel-1/2 observations; only
temperature values are used in the experiments listed in Section \ref{sec:experiments}). 

Map-like data products are available in the form of GeoTiff files, while patch-based data are stored in csv files. The different modalities of the dataset are available for download separately and can therefore be combined in a highly modular fashion. 
Download links, additional information on the dataset and useful software tools are available at \texttt{https://github.com/HSG-AIML/ben-ge}.

\section{Experimental Results}
\label{sec:experiments}

We explore the utility of the different \emph{ben-ge} data modalities and perform a range of experiments in a fully supervised setting. For this purpose, we consider the two downstream tasks of \emph{multi-label patch-based classification} and \emph{pixel-wise segmentation}.
For both tasks we utilize the extracted ESA WorldCover LULC maps as targets. We note that with regard to pixel-wise frequency of the different classes, the LULC data are highly imbalanced: 
[43.1, 1.1, 20.9, 14.1, 1.7, 0.1, 0, 18.4, 0.6, 0, 0]\% of pixels fall into the classes [tree cover, shrubland, grassland, cropland, built-up, bare/sparse vegetation, snow and ice, permanent water bodies, herbaceous wetland, mangroves, moss and lichen]. 
The classification task uses as target a one-hot encoded vector considering only those classes from the ESA WorldCover scheme that cover at least 5\% of the corresponding patch, leading to a similar class imbalance.
For the classification task we utilize a ResNet-18 model with a BCE loss function and for the segmentation task we use a U-Net model with a cross entropy loss function. 
In the case of several input data modalities, we combine these modalities in a late fusion approach: each modality is processed by a separate backbone, resulting representations are concatenated and then passed through a number of linear layers (classification) or convolutional layers (segmentation). 
In the model training process we use a Adam optimizer, no additional data augmentations and train each model for 20 epochs. Learning rates were separately chosen for each downstream task and are scheduled; the same schedule was applied to all experiments.
To evaluate our models, we make use of the F1 and accuracy metrics for the classification task and the intersection-over-union (IoU) and pixel-wise accuracy metrics for the segmentation task.
We note that, as a result of the class imbalance inherent to the dataset, the accuracy metric is highly compromised and therefore only reported for comparison purposes. 
To quantify the uncertainties inherent to our trained models, we perform each model training with 5 different random seed values and report the means and standard deviations of the resulting performances. 
We split our dataset into a training/validation/test dataset using a 0.8/0.1/0.1 split; the composition of these splits is consistent across all experiments (see Section \ref{sec:experiments_size} for details). 

We explore our dataset in the following sections with regard to the usefulness of the different Sentinel-2 channels, the dataset size and the different combinations of data modalities. In our evaluations, we will focus on the performance on the downstream tasks and computational efficiency. 

\subsection{Sentinel-2: Multispectral Data}
\label{sec:experiments_sen2}

For most \emph{BigEarthNet} or \emph{ben-ge} applications, Sentinel-2 multispectral data will form the fundamental data modality. 
However, it is unclear whether all 12 bands (Level-2A data products) contribute equally to solving the downstream tasks defined above. We therefore begin our experiments by 
training both models on three variations of Sentinel-2 data (based on the \emph{ben-ge-0.2} dataset split, see Section \ref{sec:experiments_size}): all 12 bands, only the highest resolution bands (bands 4, 3, 2 and 8 with 10~m resolution: ``RGBNIR'') and RGB (bands 4, 3 and 2),
resulting in F1 scores of 79.5\%$\pm$0.5\%, 77.1\%$\pm$0.6\% and 75.6\%$\pm$0.9\% for the classification task and IoU scores of 40.2\%$\pm$0.1\%, 39.2\%$\pm$0.1\% and 37.1\%$\pm$0.1\% for the segmentation task, respectively. 
While the combination of all 12 bands offers the best performance, the RGBNIR subset offers a good compromise, using only one third of the data and thereby providing a more efficient learning process; we adopt the RGBNIR band subset in the following experiments.

\subsection{Dataset Size}
\label{sec:experiments_size}

\begin{figure}[t!]
  \centering
  \includegraphics[width=\linewidth]{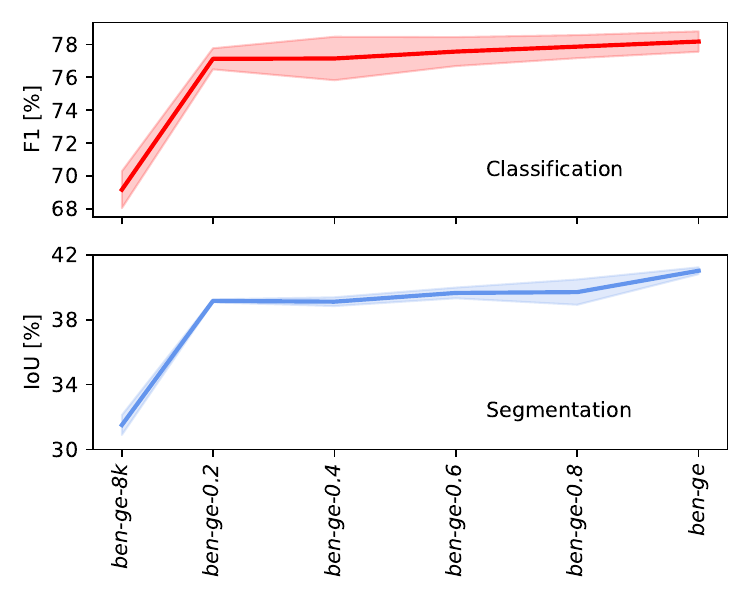}
  \caption{Visualization of model performance as a function of dataset size. We adopt the \emph{ben-ge-0.2} dataset for all our experiments in Section \ref{sec:experiments}.\label{fig:size}}
\end{figure}

\emph{ben-ge}, just like \emph{BigEarthNet}, contains 590,326 different locations and patches. We address the hypothesis that, for training the downstream tasks of classification and segmentation in a fully supervised setup, 
a smaller dataset would suffice. We therefore generate random splits of \emph{ben-ge} that contain only 20\% of the data (in the following referred to as \emph{ben-ge-0.2}, ${\sim}$118k samples), 40\% of the data (\emph{ben-ge-0.4}, ${\sim}$236k samples), 60\% of the data (\emph{ben-ge-0.6}, ${\sim}$354k samples) and 
80\% of the data (\emph{ben-ge-0.8}, ${\sim}$472k samples) across all modalities. In addition, we also create a small-scale dataset with 8k samples that are sampled in such a way as to contain the same number of samples of preferably homogeneous LULC maps per class. \emph{ben-ge-8k} is publicly available to enable quick testing on a limited, but meaningful dataset. 
While dataset splits are random, care was taken to include the smaller datasets in the larger datasets (e.g., \emph{ben-ge-0.2} is a subset of \emph{ben-ge-0.4}, which in turn is a subset of \emph{ben-ge-0.6});
the same applies to the corresponding training/validation/test splits\footnote{Index files for the individual datasets, as well as corresponding training/validation/test splits are available at \texttt{https://github.com/HSG-AIML/ben-ge}.}.
To investigate the impact of the size of the dataset on the performance of the trained model, we train models for both downstream tasks
on the training subsets of the different dataset splits. 
We find that the classification performance increases only insignificantly between 20\% of the data and the full dataset. In the case of the segmentation task, a gradual increase in IoU can be observed between 20\% of the data (37.9\%$\pm$0.7\%) and the full dataset (41.0\%$\pm$0.2\%).
Based on these findings and in order to improve the efficiency of the training process, we decide to perform all of the following experiments on the \emph{ben-ge-0.2} dataset split.

\subsection{Data Modalities}
\label{sec:experiments_modalities}

We use each of the \emph{BigEarthNet-MM} and \emph{ben-ge} data modalities, as well as different combinations thereof, as input for the training of our models. Training and evaluation are based on the \emph{ben-ge-0.2} dataset split.
Table \ref{tab:experiments_modalities} lists the resulting performances for the classification and segmentation tasks. While we use every data modality as a single input in the training, only few combinations of modalities are used as model input; 
the selection of data modality combinations is based on the performances of the 
individual modalities and thus provides only a glimpse of the possibilities. It can be observed that the impact on the performance is closely correlated to the complexity of the data modality, especially in the case of the segmentation task. 
Furthermore, we find that the classification performance increases with the number of modalities used in the process; this effect, however, is much less pronounced in the case of the segmentation task.

\begin{table*}[t]
  \caption{Model performances for different combinations of \emph{BigEarthNet-MM} and \emph{ben-ge-0.2} data modalities on the downstream tasks of multi-label patch-based classification and pixel-wise segmentation of LULC data. \label{tab:experiments_modalities}}
  \centering
  \begin{tabular}{|c|cccccc|cc|cc|}
    \hline
    N & Sen-2 & Sen-1 & Climate & DEM & Weather & Season & \multicolumn{2}{|c|}{Classification [\%]} & \multicolumn{2}{|c|}{Segmentation [\%]} \\
      &    &       &     &        &         &         &   F1-score       &  Accuracy            &    IoU      & Accuracy   \\
    \hline
    \hline
     \multirow{6}*{1} & \centering \checkmark & & & & & &                  \textbf{77.12}$\pm$0.64 & 96.21$\pm$0.08 & \textbf{39.17}$\pm$0.09 & 87.57$\pm$0.05 \\
     & & \centering \checkmark & & & & &                 73.09$\pm$0.24 & 95.60$\pm$0.05 & 31.70$\pm$0.17 & 82.65$\pm$0.05 \\
     & & & \centering \checkmark & & & &                70.50$\pm$0.34 & 94.69$\pm$0.03 & 14.70$\pm$0.32& 60.65$\pm$1.35 \\
     & & & & \centering \checkmark & & &                 55.96$\pm$1.00 & 93.53$\pm$0.15 & 26.25$\pm$0.48 & 76.92$\pm$0.63\\     
     & & & & & \centering \checkmark & &                 46.15$\pm$0.68 & 91.60$\pm$0.02 & 6.30$\pm$0.05 & 45.20$\pm$0.08\\
     & & & & & & \centering \checkmark &                 39.15$\pm$0.74 & 91.75$\pm$0.05 & 6.01$\pm$0.34 & 43.89$\pm$0.51\\
     \hline
     \multirow{2}*{2} & \centering \checkmark & \centering \checkmark & & & & &  \textbf{82.81}$\pm$0.29 & 97.03$\pm$0.04  & \textbf{39.67}$\pm$0.16 & 87.98$\pm$0.07 \\
     & \centering \checkmark & & & & & \centering \checkmark &  78.61$\pm$0.67 & 96.42$\pm$0.08 & 38.92$\pm$0.21 & 87.37$\pm$0.10\\
     \hline
     \multirow{3}*{3} & \centering \checkmark & \centering \checkmark & \centering \checkmark & & & &  \textbf{85.12}$\pm$0.34 & 97.39$\pm$0.05  & 39.63$\pm$0.23 & 87.94$\pm$0.12 \\
     & \centering \checkmark & \centering \checkmark & & \centering \checkmark & & &  83.30$\pm$0.43 & 97.10$\pm$0.08 & \textbf{39.71}$\pm$0.21 & 88.05$\pm$0.11\\
     & \centering \checkmark & \centering \checkmark & & & & \centering \checkmark & --- & --- & 39.61$\pm$0.19 & 87.93$\pm$0.12 \\
     \hline
  \end{tabular}

\end{table*}

\section{Discussion and Conclusion}

Our experiments provide insights into the capabilities and usefulness of the \emph{ben-ge} dataset and data fusion across a number of data modalities. 
Results from Sections \ref{sec:experiments_sen2} and \ref{sec:experiments_size} indicate that there is only a moderate
benefit to using more than the RGBNIR subset of the Sentinel-2 bands or more than 20\% of the dataset in a fully supervised training scenario for the LULC classification task. The comparison with the segmentation task results indicate different effects for other downstream
tasks, other targets or other data modalities or combinations thereof. 
Results reported in Table \ref{tab:experiments_modalities} show the varying value of the different data modalities for the downstream tasks defined in Section \ref{sec:experiments}.
In general, it can be observed that more complex (map-like) data types contribute stronger to the model performance, especially in the case of the segmentation task. 
Interestingly, patch-based climate zone information provides rather strong constraints for the classification task. Furthermore, it seems that the overall performance improves with the number of data modalities.

Our results provide a glimpse of the usefulness of extending Earth observation datasets across a range of data modalities. By relying only on freely and globally available data products, the data modalities presented here can be generated for and 
therefore utilized to enhance any other Earth observation dataset, offering new opportunities to improve the performance of deep learning models for Earth observation applications in general and self-supervised model pretraining in particular.


\vspace{2em}

This work is funded by Swiss National Science Foundation research project grant 213064.

\bibliographystyle{IEEEbib}
\bibliography{refs}

\end{document}